\documentclass[conference]{IEEEtran}
\IEEEoverridecommandlockouts

\usepackage{amsmath,amssymb,amsfonts}
\usepackage{graphicx}
\usepackage{booktabs}
\usepackage{multirow}
\usepackage{siunitx}
\usepackage{xcolor}
\usepackage[hidelinks]{hyperref}
\usepackage{cite}

\newcommand{\vect}[1]{\boldsymbol{#1}}
\newcommand{\R}{\mathbb{R}}
\DeclareMathOperator{\diag}{diag}

\begin{document}

\title{An Open MuJoCo Simulation Framework for a Quadrotor Carrying a Tendon-Driven Soft Continuum Arm}

\author{\IEEEauthorblockN{Haitham El-Hussieny}
\IEEEauthorblockA{Egypt-Japan University of Science and Technology (E-JUST)\\
Alexandria, Egypt\\
haitham.elhussieny@ejust.edu.eg}}

\maketitle

\begin{abstract}
Combining aerial robots with soft continuum manipulators promises safe, compliant interaction with the environment, but the coupled dynamics of a free-flying base and a highly deformable, underactuated arm are difficult to study without a suitable simulator. This paper describes an open-source simulation framework, built on the MuJoCo physics engine, for a quadrotor that carries a tendon-driven soft continuum arm underneath its body. The arm is modelled as a pseudo-rigid-body chain of twelve segments joined by pairs of orthogonal compliant hinges. Three tendons, spaced \SI{120}{\degree} apart and routed through every segment, bend it in any direction. A geometric tracking controller on $SE(3)$ flies the vehicle; its rotor allocation matrix is generated directly from the model description. A closed-form tendon-tension allocation keeps every tension within its limits and gives the same bending moment in every direction. We report the full set of modelling choices and parameters, and we evaluate the framework in four simulated scenarios: a \SI{36}{\second} mission with a figure-eight and a tendon-driven tip sweep, an attitude-gain study, a quasi-static characterization of the bending workspace, and an impulsive disturbance applied to the arm tip. With the nominal gains the vehicle tracks the mission with \SI{3.1}{\centi\meter} RMS position error while the arm tip swings up to \SI{37}{\centi\meter} from the body axis. The gain study shows that a low-bandwidth attitude loop lets the arm excite sustained oscillations. The simulation runs about $15\times$ faster than real time on a laptop CPU. All models, controllers and scripts that reproduce the figures are released.
\end{abstract}


\section{Introduction}
Aerial manipulators extend the workspace of robotic arms to places that ground robots cannot reach, such as infrastructure at height, canopies and disaster sites~\cite{ruggiero2018aerial}. Most aerial manipulators use rigid, serial arms. Rigid arms are precise, but they transmit contact forces straight to the flying base and can be unsafe near people and fragile objects. Soft and continuum robots~\cite{rus2015design,kim2013soft,webster2010design} offer an attractive alternative. Their intrinsic compliance absorbs impacts, adapts to object geometry, and makes the arm safer to operate near people. Soft grippers have already been used on drones for dynamic grasping~\cite{fishman2021dynamic}.

Attaching a soft continuum arm to a multirotor raises its own modelling and control problems. The arm has many lightly damped degrees of freedom. Its motion shifts the centre of mass of the whole system and exerts reaction wrenches on the base. Gravity, inertial loads from flight manoeuvres and the actuation all deform it at the same time. Quadrotor tracking controllers, from geometric control on $SE(3)$~\cite{lee2010geometric} to data-driven approaches such as deep Koopman-based model predictive control~\cite{el2026real}, are usually designed and evaluated for a rigid vehicle, and a swinging soft arm violates that assumption. Before any hardware is built, designers need a simulator that captures these couplings well enough to choose arm stiffness, tendon routing, actuator limits and controller gains. General-purpose multirotor simulators such as Gazebo with RotorS~\cite{koenig2004gazebo,furrer2016rotors} focus on rigid vehicles and rigid payloads. Detailed continuum-robot models based on Cosserat rods~\cite{rucker2011statics} are usually solved on a fixed base and are too expensive to run in closed loop with flight dynamics.

This paper documents a lightweight framework that fills this gap using MuJoCo~\cite{todorov2012mujoco}, a fast physics engine for articulated systems with contact. The contributions are:
\begin{enumerate}
  \item An MJCF model of an X-configuration quadrotor that carries a tendon-driven soft continuum arm. The arm uses a pseudo-rigid-body discretization with two-axis compliant joints and three spatial tendons (Section~\ref{sec:model}).
  \item A flight controller that combines a geometric $SE(3)$ tracking law with an allocation matrix derived from the model's actuator definitions. It is paired with a closed-form, saturation-free tendon allocation that bends the arm equally in every direction (Section~\ref{sec:control}).
  \item A set of reproducible simulation experiments that quantify tracking performance, the effect of attitude-loop bandwidth on arm-induced oscillations, the bending workspace, disturbance rejection and computational cost (Section~\ref{sec:results}).
\end{enumerate}
The code is publicly available (see Code Availability). Every number and figure in this paper is produced by a single script.

\section{System Modelling}\label{sec:model}
\begin{figure*}[t]
  \centering
  \includegraphics[width=\textwidth]{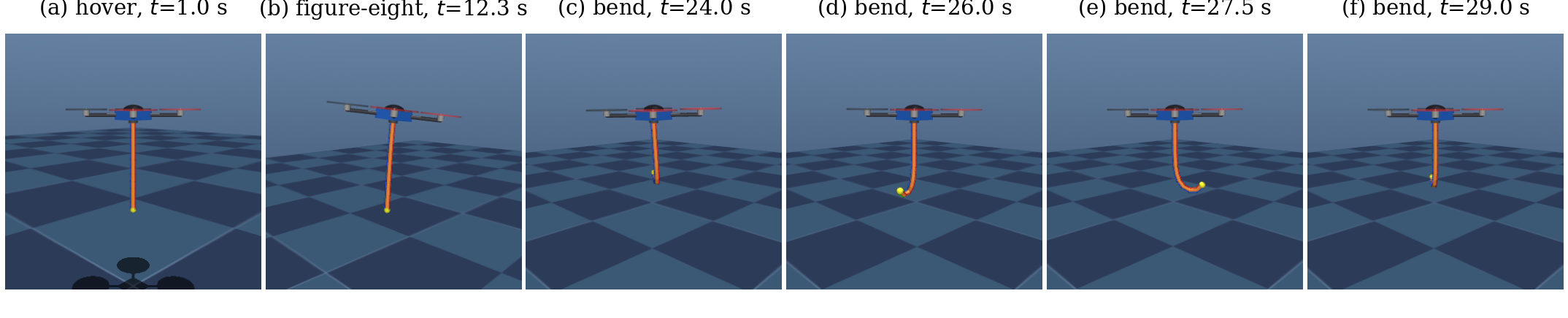}
  \caption{Rendered frames from the demonstration mission. (a)~Hover with the arm hanging under gravity. (b)~Figure-eight flight with the arm deflected by inertial loads. (c)--(f)~Hover while the tendons sweep the bending direction through a full revolution. The arm curls into a hook toward the tensioned side.}
  \label{fig:snapshots}
\end{figure*}

Fig.~\ref{fig:snapshots} shows the simulated system. Table~\ref{tab:params} lists its parameters. The model is written in MuJoCo's MJCF format and compiled by MuJoCo~3.15. It has $n_q=31$ position coordinates and $n_v=30$ velocity coordinates: a 6-DoF free joint for the quadrotor and 24 hinge joints for the arm. There are seven actuators (four rotors and three tendons).

\begin{table}[t]
\centering
\caption{Model parameters}
\label{tab:params}
\begin{tabular}{@{}llr@{}}
\toprule
& Parameter & Value \\
\midrule
\multirow{6}{*}{\rotatebox{90}{Quadrotor}}
& Mass (frame, rotors, flange) & \SI{1.06}{\kilo\gram} \\
& Principal inertia & $(6.7, 6.7, 12.9)\times10^{-3}$\,\si{\kilo\gram\meter\squared} \\
& Arm (rotor) offset $(\pm x_r, \pm y_r)$ & $(\pm0.177, \pm0.177)$\,\si{\meter} \\
& Thrust limit per rotor & \SIrange{0}{8}{\newton} \\
& Drag-to-thrust ratio $\kappa$ & \SI{0.016}{\meter} \\
& Hover thrust per rotor & \SI{3.09}{\newton} \\
\midrule
\multirow{10}{*}{\rotatebox{90}{Continuum arm}}
& Segments $N$ / segment length $\ell$ & 12 / \SI{40}{\milli\meter} \\
& Segment radius / mass & \SI{12}{\milli\meter} / \SI{15}{\gram} \\
& Tip mass (end-effector sphere) & \SI{20}{\gram} \\
& Arm mass / backbone length & \SI{0.20}{\kilo\gram} / \SI{0.48}{\meter} \\
& Tip depth below hub centre & \SI{0.53}{\meter} \\
& Joint stiffness $k$ (each axis) & \SI{0.06}{\newton\meter\per\radian} \\
& Joint damping $c$ (each axis) & \SI{0.01}{\newton\meter\second\per\radian} \\
& Joint range & $\pm\SI{0.6}{\radian}$ \\
& Tendon offset $r$ / angles $\phi_i$ & \SI{16}{\milli\meter} / $0,120,240^\circ$ \\
& Tendon tension limit $T_{\max}$ & \SIrange{0}{6}{\newton} \\
\midrule
\multirow{3}{*}{\rotatebox{90}{Sim.}}
& Time step / integrator & \SI{2}{\milli\second} / \texttt{implicitfast} \\
& Controller rate & \SI{500}{\hertz} (every step) \\
& Air density / viscosity & \SI{1.2}{\kilo\gram\per\cubic\meter} / \SI{1.8e-5}{\pascal\second} \\
\bottomrule
\end{tabular}
\end{table}

\subsection{Quadrotor}
The quadrotor is a single rigid body attached to the world by a free joint. Its position is $\vect{p}\in\R^3$, its attitude is $R\in SO(3)$, its linear velocity in the world frame is $\vect{v}$, and its angular velocity in the body frame is $\vect{\omega}$. Each rotor $j\in\{1,\dots,4\}$ sits at $(x_j, y_j)$ in the body frame and is modelled as a MuJoCo \texttt{motor} acting on a site at the rotor hub. Its gear vector is $[0,0,1,0,0,s_j\kappa]^\top$, so a control input $u_j$ (in newtons) applies a thrust $u_j$ along the body $z$-axis and a reaction (drag) torque $s_j\kappa u_j$ about it. The spin directions alternate, $s_j\in\{+1,-1,+1,-1\}$. The resulting body wrench is
\begin{equation}
\begin{bmatrix} f \\ \vect{\tau} \end{bmatrix}
= A\,\vect{u},\quad
A = \begin{bmatrix}
1 & 1 & 1 & 1\\
y_1 & y_2 & y_3 & y_4\\
-x_1 & -x_2 & -x_3 & -x_4\\
s_1\kappa & s_2\kappa & s_3\kappa & s_4\kappa
\end{bmatrix},
\label{eq:alloc}
\end{equation}
where $f$ is the collective thrust and $\vect{\tau}$ the body torque. The controller builds $A$ at start-up from the compiled model (site positions and actuator gears) rather than from hand-coded constants. Changing the airframe in the MJCF file therefore cannot desynchronize the mixer. Rotor speed dynamics, motor lag and blade-element aerodynamics are deliberately left out (Section~\ref{sec:discussion}).

\subsection{Soft continuum arm}
Simulating the arm as a Cosserat rod would require a dedicated solver. We instead use a pseudo-rigid-body (PRB) discretization~\cite{howell2001compliant}, which MuJoCo's articulated-body machinery handles efficiently. The arm consists of $N=12$ capsules of length $\ell$, connected in series below a mounting flange on the quadrotor. Each link $k$ connects to its parent through two orthogonal hinges, about the local $x$ and $y$ axes, with angles $\vect{\theta}_k=(\theta_{k,x},\theta_{k,y})$. Each hinge has a linear spring of stiffness $k$ and a viscous damper $c$. The elastic potential energy is therefore
\begin{equation}
V_e = \tfrac{1}{2}\,k\sum_{k=1}^{N}\|\vect{\theta}_k\|^2 ,
\end{equation}
which is the discretized bending energy $\tfrac12\int EI\,\kappa^2\,ds$ of a rod with flexural rigidity $EI = k\ell$. With the values of Table~\ref{tab:params}, $EI=\SI{2.4e-3}{\newton\meter\squared}$. For a solid circular cross-section of radius \SI{12}{\milli\meter}, this corresponds to a Young's modulus of about \SI{0.15}{\mega\pascal}, which is within the range of soft silicone elastomers. Torsion and axial extension are not modelled. The \SI{20}{\gram} sphere at the distal end stands in for a small gripper or sensor.

\emph{Contacts.} The arm geoms use a collision bitmask that lets them touch the ground and the quadrotor frame but not each other. Neighbouring links overlap at their capsule caps, so self-collisions would otherwise produce spurious contact forces.

\subsection{Tendon actuation}
Three spatial tendons $i\in\{a,b,c\}$ run parallel to the backbone at a radial offset $r$ and angular positions $\phi_i\in\{0^\circ,120^\circ,240^\circ\}$. Each tendon is anchored on the flange and passes through a via-point (MuJoCo \emph{site}) at the distal end of every link. MuJoCo computes each tendon's length $L_i(\vect{q})$ as the sum of the straight-line distances between consecutive sites, together with the moment arms $\partial L_i/\partial\vect{q}$. A \texttt{motor} actuator with gear $-1$ converts a non-negative control $T_i\in[0,T_{\max}]$ into a tension that shortens the tendon. The generalized force is $\vect{\tau}_q=-\sum_i T_i\,(\partial L_i/\partial\vect{q})^\top$.

For a straight arm, the forces at the intermediate via-points cancel. A tensioned tendon then produces the same bending moment of magnitude $rT_i$ at every joint, directed about the axis $(\sin\phi_i,-\cos\phi_i,0)$. In the absence of gravity this produces approximately constant curvature, the usual idealization of tendon-driven continuum robots~\cite{webster2010design}. The net bending moment is
\begin{equation}
\vect{m} = r\sum_{i} T_i\,\vect{n}_i,\qquad \vect{n}_i=(\cos\phi_i,\sin\phi_i)^\top,
\label{eq:moment}
\end{equation}
where $\vect{m}$ is expressed by the in-plane direction toward which the arm bends. Gravity adds a configuration-dependent restoring moment. In our parameter range it dominates the elastic stiffness, so pulling a tendon curls the distal part of the arm into a hook rather than a uniform arc (Fig.~\ref{fig:snapshots}c--f). The reaction to the tendon tensions acts on the flange, so the quadrotor experiences the internal loads of the arm.

\subsection{Environment and integration}
MuJoCo's built-in inertia-based fluid model is enabled with the density and viscosity of air. It adds drag to every body, which provides a mild amount of aerodynamic damping for the swinging arm. Rotor downwash on the arm is not modelled. The equations of motion are integrated with MuJoCo's \texttt{implicitfast} integrator, which treats the joint damping implicitly, at a time step of \SI{2}{\milli\second}.

\section{Control}\label{sec:control}
The controller has full state access and runs at every simulation step. It treats the arm as an unmodelled disturbance and uses only the total mass $m_\Sigma=\SI{1.26}{\kilo\gram}$ of the quadrotor and arm. This deliberately simple baseline isolates the effect of the soft arm on a standard flight controller.

\subsection{Position loop}
Given a reference $(\vect{p}_d,\vect{v}_d,\vect{a}_d,\psi_d)$, the commanded acceleration is
\begin{equation}
\vect{a}_c = K_p\vect{e}_p + K_d\vect{e}_v + K_i\!\int\!\vect{e}_p\,dt + \vect{a}_d,
\end{equation}
with $\vect{e}_p=\vect{p}_d-\vect{p}$, $\vect{e}_v=\vect{v}_d-\vect{v}$, and an integrator clamped to $\pm\SI{1}{\meter\second}$. The horizontal part of $\vect{a}_c$ is saturated so that the commanded tilt never exceeds \SI{35}{\degree}. The desired force is $\vect{F}_d=m_\Sigma(\vect{a}_c+g\vect{e}_3)$.

\subsection{Attitude loop}
Following~\cite{lee2010geometric}, the desired body axes are $\vect{b}_{3d}=\vect{F}_d/\|\vect{F}_d\|$, $\vect{b}_{2d}=\vect{b}_{3d}\times\vect{b}_{1c}/\|\vect{b}_{3d}\times\vect{b}_{1c}\|$ with $\vect{b}_{1c}=(\cos\psi_d,\sin\psi_d,0)$, and $\vect{b}_{1d}=\vect{b}_{2d}\times\vect{b}_{3d}$. Together they give $R_d=[\vect{b}_{1d}\;\vect{b}_{2d}\;\vect{b}_{3d}]$. The thrust and torque commands are
\begin{align}
f &= \vect{F}_d^\top R\,\vect{e}_3,\\
\vect{e}_R &= \tfrac12\left(R_d^\top R-R^\top R_d\right)^\vee,\\
\vect{\tau} &= -K_R\,\vect{e}_R - K_\omega\,\vect{\omega},
\end{align}
where $(\cdot)^\vee$ maps a skew-symmetric matrix to a vector. We omit the feed-forward angular velocity and gyroscopic terms of the full law because the references used here are slow relative to the attitude dynamics. The rotor commands are $\vect{u}=\mathrm{sat}_{[0,8]}\big(A^{-1}[f;\vect{\tau}]\big)$. The nominal gains are $K_p=\diag(6,6,10)$, $K_d=\diag(4,4,6)$, $K_i=\diag(1,1,2)$, $K_R=\diag(4,4,0.8)$ and $K_\omega=\diag(0.4,0.4,0.15)$ in SI units. Section~\ref{sec:gains} justifies the attitude gains.

\subsection{Tendon tension allocation}
The bending command is a direction $\beta$ in the body $xy$-plane and a normalized magnitude $\mu\in[0,1]$. Because tendons can only pull, the tensions must satisfy $0\le T_i\le T_{\max}$. Let $c_i=\cos(\beta-\phi_i)$. For three equally spaced tendons, $\sum_i\vect{n}_i\vect{n}_i^\top=\tfrac32 I$ and $\sum_i\vect{n}_i=\vect{0}$. Hence the minimum-norm solution of~\eqref{eq:moment}, $T_i\propto c_i$, can be shifted by any common-mode tension without changing the net moment. We choose
\begin{equation}
T_i = \frac{\mu\,T_{\max}}{\sqrt3}\,\Big(c_i-\min_j c_j\Big),
\label{eq:tension}
\end{equation}
which sets the slackest tendon to zero. Over all $\beta$, the spread $\max_j c_j-\min_j c_j$ reaches its largest value, $\sqrt3$, when $\beta$ lies midway between two tendons. Therefore~\eqref{eq:tension} never exceeds $T_{\max}$, and no clipping is needed. Substituting into~\eqref{eq:moment} gives
\begin{equation}
\vect{m}=\tfrac{\sqrt3}{2}\,r\,\mu\,T_{\max}\,(\cos\beta,\sin\beta)^\top
\end{equation}
for every $\beta$. The bending authority is thus direction-independent. We verified numerically that the peak tension is exactly $T_{\max}$ and that the resultant is constant at $\tfrac{\sqrt3}{2}T_{\max}=\SI{5.20}{\newton}$ over a $0.1^\circ$ sweep of $\beta$. An earlier version of the framework used $T_i=\mu T_{\max}\max(c_i,0)$. That map loses half of its authority between tendons, and in simulation it produced a tip workspace whose radius varied between \SI{10.7}{} and \SI{15.9}{\centi\meter} with direction.

\subsection{Reference generation}
Transitions use the minimum-jerk blend $\sigma(s)=10s^3-15s^4+6s^5$~\cite{flash1985coordination}. The horizontal figure-eight is $\vect{p}_d=(A\sin\Omega\tau,\;0.6A\sin2\Omega\tau,\;z_c)$ with $\Omega=2\pi/8~\si{\radian\per\second}$ and $z_c=\SI{1.6}{\meter}$. The amplitude $A$ is ramped from 0 to \SI{1.2}{\meter} and back with $\sigma$ over \SI{2}{\second} at each end of the segment. Velocity and acceleration feed-forward terms are obtained by central finite differences of the reference. The peak reference speed is \SI{1.47}{\meter\per\second}, and the peak horizontal acceleration is \SI{3.2}{\meter\per\second\squared}, equivalent to a quasi-static tilt of \SI{18}{\degree}.

\section{Simulation Experiments}\label{sec:results}
All experiments were run with MuJoCo~3.15.0 and Python~3.12 on an Apple~M3~Pro laptop. They are generated by \texttt{paper/make\_figures.py}. The vehicle always starts at rest at \SI{1}{\meter} altitude with the arm hanging straight.

\subsection{Demonstration mission}
\begin{figure}[t]
  \centering
  \includegraphics[width=\columnwidth]{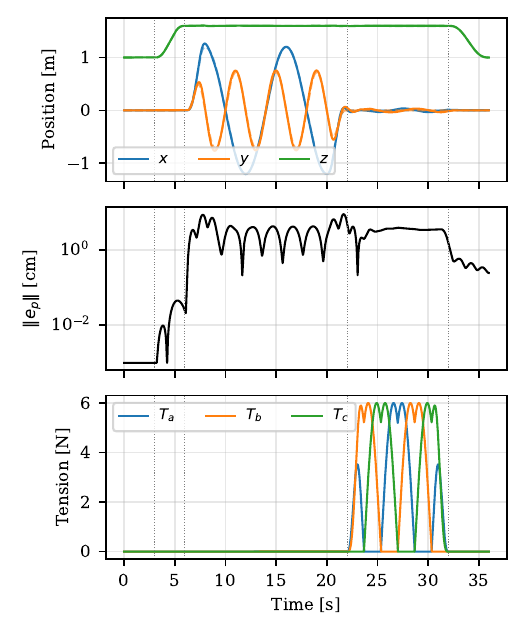}
  \caption{Demonstration mission. Top: quadrotor position (solid) and reference (dashed). Middle: position error norm (log scale). Bottom: tendon tensions from~\eqref{eq:tension}. Dotted lines mark the phase boundaries.}
  \label{fig:mission}
\end{figure}

\begin{figure}[t]
  \centering
  \includegraphics[width=\columnwidth]{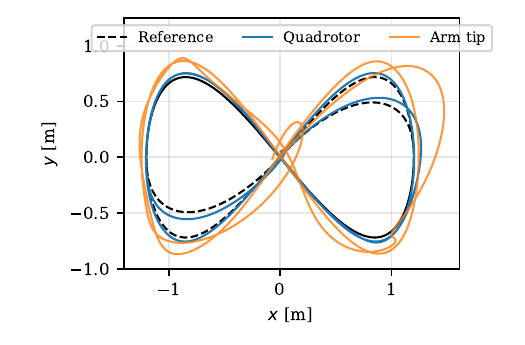}
  \caption{Top view of the figure-eight phase. The arm tip lags the vehicle and swings outward at the turns.}
  \label{fig:topview}
\end{figure}

The \SI{36}{\second} mission consists of five phases: (i)~hover at \SI{1.0}{\meter}; (ii)~climb to \SI{1.6}{\meter}; (iii)~a \SI{16}{\second} figure-eight with the tendons slack; (iv)~a \SI{10}{\second} hover during which the bending direction rotates at 0.2 revolutions per second and $\mu$ is ramped to~1; and (v)~a return to the start point. Fig.~\ref{fig:mission} shows the tracking result, Fig.~\ref{fig:topview} the horizontal path, and Table~\ref{tab:phases} the statistics for each phase.

\begin{table}[t]
\centering
\caption{Mission statistics per phase (nominal gains)}
\label{tab:phases}
\setlength{\tabcolsep}{4pt}
\begin{tabular}{@{}lrrrrc@{}}
\toprule
Phase & \multicolumn{2}{c}{$\|\vect{e}_p\|$ [cm]} & Tilt & Tip lat. & Thrust \\
 & RMS & max & max [$^\circ$] & max [cm] & range [N] \\
\midrule
Hover        & $<$0.01 & $<$0.01 & 0.0  & 0.0  & 3.09--3.09 \\
Climb        & 0.03 & 0.05 & 0.0  & 0.0  & 2.97--3.21 \\
Figure-eight & 3.90 & 8.87 & 23.5 & 36.8 & 2.95--3.53 \\
Tendon sweep & 3.34 & 5.16 & 4.7  & 21.8 & 2.73--3.42 \\
Return       & 0.50 & 1.50 & 0.6  & 2.1  & 3.02--3.16 \\
\midrule
Whole mission & 3.14 & 8.87 & 23.5 & 36.8 & 2.73--3.53 \\
\bottomrule
\end{tabular}
\end{table}

The largest errors occur during the figure-eight. Here the arm, whose mass is \SI{16}{\percent} of the total, lags behind the vehicle (Fig.~\ref{fig:topview}) and swings laterally by up to \SI{37}{\centi\meter} from the body axis. The reaction to this swing is a disturbance that the controller does not model. The thrusts stay between \SI{2.7}{} and \SI{3.6}{\newton}, well within the \SI{8}{\newton} limit. The thrust-to-weight ratio is 2.6, so there is ample margin.

During the tendon sweep the position error rises to \SI{3.3}{\centi\meter} RMS even though the vehicle is commanded to hover. Static tests (Section~\ref{sec:bending}) explain why. When the arm bends, its weight acts off the body axis and creates a pitching or rolling moment. The attitude loop is a PD law without integral action, so it balances this moment with a small steady attitude error. With the nominal gains this error is only about \SI{0.02}{\degree}, but the horizontal thrust component it produces makes the vehicle drift slowly toward the bending side until the position integrator compensates. In a held-bend test at $\mu=1$ the drift was \SI{1.9}{\centi\meter} after \SI{5}{\second}, \SI{0.3}{\centi\meter} after \SI{15}{\second} and \SI{0.02}{\centi\meter} after \SI{30}{\second}. A feed-forward term based on the arm's centre of mass would remove most of this error.

\subsection{Effect of attitude-loop bandwidth}\label{sec:gains}
Our first implementation used the soft attitude gains listed first in Table~\ref{tab:gains}. With these gains the vehicle never settled during hover; it kept oscillating together with the arm. To quantify this effect we repeated the mission with four attitude gain sets, keeping the position gains fixed (Table~\ref{tab:gains}). With $J\approx\SI{6.7e-3}{\kilo\gram\meter\squared}$, the roll/pitch natural frequency $\sqrt{K_R/J}$ of the bare vehicle ranges from about \SI{7}{} to \SI{24}{\radian\per\second} across the four sets. With the softest gains, a sustained, low-frequency (below \SI{0.5}{\hertz}) coupled oscillation between the vehicle and the arm persists through the hover phases. Its RMS error after $t=\SI{24}{\second}$ is \SI{34}{\centi\meter}, larger than during the aggressive figure-eight. Each doubling of $K_R$ roughly halves the error. The nominal set, the stiffest one tested, still leaves a wide thrust margin and reduces the RMS error to \SI{3.1}{\centi\meter}. The attitude loop therefore needs a bandwidth well above the dominant frequencies of the arm. This requirement is worth checking against the real motor and ESC bandwidth when the design is transferred to hardware.

\begin{table}[t]
\centering
\caption{Mission performance versus roll/pitch attitude gains}
\label{tab:gains}
\setlength{\tabcolsep}{3.5pt}
\begin{tabular}{@{}lccrrrr@{}}
\toprule
Set & $K_{R,xy}$ & $K_{\omega,xy}$ & RMS & max & RMS$_{t>24\,\mathrm{s}}$ & Tilt$_{\max}$ \\
 & & & [cm] & [cm] & [cm] & [$^\circ$] \\
\midrule
Soft    & 0.35 & 0.06 & 22.67 & 52.82 & 33.59 & 40.7 \\
Medium  & 1.0  & 0.15 & 7.99  & 16.70 & 11.04 & 28.6 \\
Stiff   & 2.0  & 0.25 & 4.80  & 11.97 & 5.56  & 25.4 \\
Nominal & 4.0  & 0.40 & 3.14  & 8.87  & 2.85  & 23.5 \\
\bottomrule
\end{tabular}
\\[2pt]{\footnotesize Yaw gains $(K_{R,z},K_{\omega,z})$: (0.1,\,0.04), (0.3,\,0.08), (0.5,\,0.1), (0.8,\,0.15).}
\end{table}

\subsection{Bending workspace}\label{sec:bending}
\begin{figure}[t]
  \centering
  \includegraphics[width=\columnwidth]{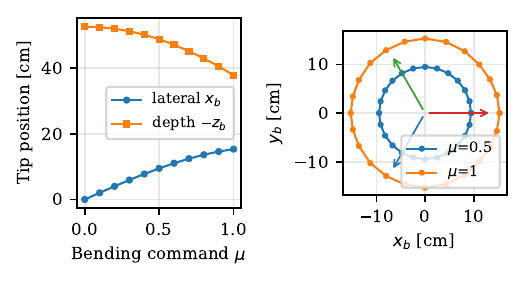}
  \caption{Quasi-static bending in hover. Left: tip lateral deflection and depth versus the bending command $\mu$ for $\beta=0$. Right: tip workspace in the body $xy$-plane when $\beta$ is swept, for $\mu=0.5$ and $\mu=1$. Arrows show the tendon directions $\phi_i$.}
  \label{fig:bending}
\end{figure}

To characterize the arm independently of the flight manoeuvres, the vehicle is commanded to hover at \SI{1.6}{\meter}. A constant bending command is ramped in over \SI{1}{\second}, and the tip position, expressed in the body frame, is recorded after \SI{5}{\second}. Fig.~\ref{fig:bending} (left) shows the tip position for $\beta=0$ as $\mu$ increases. The lateral deflection grows almost linearly at first and saturates at \SI{15.3}{\centi\meter} at $\mu=1$. Meanwhile the tip rises from \SI{52.6}{} to \SI{37.8}{\centi\meter} below the body, because the distal links curl back. The vehicle's tilt stays below \SI{0.03}{\degree} throughout. Fig.~\ref{fig:bending} (right) sweeps $\beta$ in \SI{15}{\degree} steps. Thanks to~\eqref{eq:tension}, the tip traces a nearly circular workspace whose radius varies only between \SI{14.95}{} and \SI{15.32}{\centi\meter} ($\pm1.2\%$) at $\mu=1$. The remaining anisotropy comes from the serial $x$--$y$ hinge order and the discrete tendon routing.

\subsection{Disturbance on the arm tip}
\begin{figure}[t]
  \centering
  \includegraphics[width=\columnwidth]{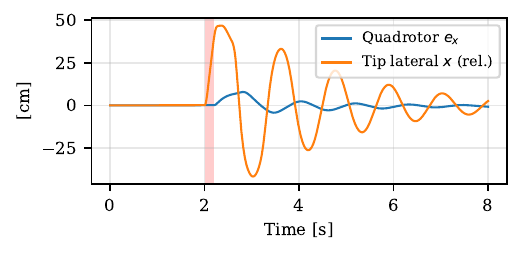}
  \caption{Response to a \SI{0.4}{\newton\second} lateral impulse on the arm tip (shaded) during hover: vehicle position error and tip lateral displacement relative to the vehicle.}
  \label{fig:disturbance}
\end{figure}

With the vehicle hovering at \SI{1}{\meter}, a lateral force of \SI{2}{\newton} is applied to the distal link for \SI{0.2}{\second}, an impulse of \SI{0.4}{\newton\second} (Fig.~\ref{fig:disturbance}). The tip swings by up to \SI{47}{\centi\meter} relative to the body. The vehicle tilts by at most \SI{7.0}{\degree} and deviates by at most \SI{7.9}{\centi\meter}. Its position returns to within \SI{1}{\centi\meter} of the set-point \SI{5.0}{\second} after the push. The lightly damped swing of the arm, rather than the flight controller, determines this recovery time. Faster recovery would require more structural damping in the arm or active damping through the tendons.

\subsection{Computational cost}
The full mission ($18\,000$ steps, including the Python controller and reference generation at every step) takes \SI{2.3}{\second} of wall-clock time, or about \SI{129}{\micro\second} per step. This is a real-time factor of about $15$ on a single CPU core without rendering. The framework is therefore fast enough for interactive use, for gain sweeps and Monte-Carlo studies, and as a candidate environment for learning-based controllers.

\section{Discussion and Limitations}\label{sec:discussion}
The framework is intended for early design studies and controller prototyping, not for high-fidelity prediction. Its main limitations are:

\emph{Arm model.} The PRB chain captures bending, gravity loading, inertia and tendon actuation. It neglects shear, torsion, axial extension, material nonlinearity, hysteresis and tendon friction. Its parameters are representative but have not been identified from a physical prototype. Comparing it with a Cosserat-rod model~\cite{rucker2011statics} and with experiments is a natural next step. Increasing $N$ refines the discretization at a linear cost in computation.

\emph{Aerodynamics.} Rotors are ideal force generators without speed dynamics. Downwash, ground effect and rotor--arm aerodynamic interaction are not modelled. Downwash could noticeably load a light, slender arm hanging directly below the rotors.

\emph{Sensing and control.} The controller uses exact state feedback and treats the arm as a disturbance. A rigorous treatment of the coupled system calls for estimation of the arm's shape and model-based control that accounts for the arm~\cite{dellasantina2023model}. Examples include feed-forward compensation of the centre of mass or of the reaction wrench, and coordinated planning of the vehicle and the arm. The model makes these extensions straightforward to prototype.

\section{Conclusion}
We presented an open MuJoCo framework for simulating a quadrotor that carries a tendon-driven soft continuum arm. It comprises a PRB arm with spatial tendons, a geometric flight controller whose rotor allocation comes from the model, and a closed-form tendon allocation that is saturation-free and bends the arm equally in every direction. Reproducible experiments showed centimetre-level tracking with \SI{3.1}{\centi\meter} RMS error during aggressive manoeuvres. They identified attitude-loop bandwidth as a key requirement when flying with a soft arm, and they characterized a bending workspace of about \SI{15}{\centi\meter} radius that is nearly identical in every direction. The simulation runs about $15\times$ faster than real time. Future work will add motor dynamics and a downwash model, compensation that accounts for the arm's dynamics, and validation against a physical prototype.

\section*{Code Availability}
The MJCF model (\texttt{quad\_continuum.xml}), the controller (\texttt{controller.py}), the interactive demonstration (\texttt{demo.py}), and the script that regenerates every figure and number in this paper (\texttt{paper/make\_figures.py}) are available at \url{https://github.com/elhussieny/quad-soft-continuum}.

\bibliographystyle{IEEEtran}
\bibliography{refs}

\end{document}